\documentclass[letterpaper, 10 pt, conference]{ieeeconf}
\IEEEoverridecommandlockouts              
\usepackage{epsfig} 
\usepackage{mathptmx} 
\usepackage{amsmath, amsfonts} 
\usepackage{array}
\usepackage{amssymb}  
\usepackage{bm}
\usepackage{verbatim}
\usepackage{stfloats}
\usepackage{siunitx}
\usepackage{graphicx}
\graphicspath{{figures/}}
\usepackage{tablefootnote}
\usepackage{subcaption}
\usepackage{hyperref}
\usepackage{tabularx}
\usepackage{mathtools}
\usepackage{wrapfig}
\usepackage{derivative}
\usepackage{lipsum}
\usepackage{pgfplots}
\pgfplotsset{width=10cm,compat=1.9}
\usepgfplotslibrary{external}
\usepackage{subfiles}
\usepackage{amsfonts}
\usepackage{textcomp}
\usepackage[all]{nowidow}
\DeclareUnicodeCharacter{221E}{\ensuremath{_\infty}}  
\usepackage{algorithm}
\usepackage{algpseudocode}
\usepackage{cleveref}
\usepackage{xpatch}
\usepackage{pgf}
\usepackage{import}
\usepackage{diagbox}
\usepackage{balance}
\usepackage{booktabs}
\usepackage{pifont}
\usepackage[acronym]{glossaries}
\usepackage{placeins}
\DeclareMathAlphabet{\mathcal}{OMS}{cmsy}{m}{n}

\newacronym{esu}{ESU}{electrosurgical unit}
\newacronym{pdf}{PDF}{probability density function}
\newacronym{tukf}{TUKF}{``Truncated Unscented Kalman Filter”}
\newacronym{thermo}{ThERMO}{Thermography for Electrosurgical Rate Modulation via Optimization}
\newacronym{star}{STAR}{Smart Tissue Autonomous Robot}
\newacronym{ir}{IR}{infrared}
\newacronym{pva}{PVA}{polyvinyl alcohol}

\title{\LARGE \bf Towards Autonomous Robotic Electrosurgery via Thermal Imaging}
\author{Naveed D. Riaziat, Joseph Chen, Axel Krieger,~\IEEEmembership{Senior Member,~IEEE,}\\ Jeremy D. Brown, ~\IEEEmembership{Senior Member,~IEEE}
\thanks{Funding for this project was provided by the Link Foundation Modeling, Simulation, and Training Fellowship and ARPA-H grants AY1AX000023 and D24AC00415.}
\thanks{© 2025 IEEE. Personal use of this material is permitted. Permission from IEEE must be obtained for all other uses, in any current or future media, including reprinting/republishing this material for advertising or promotional purposes, creating new collective works, for resale or redistribution to servers or lists, or reuse of any copyrighted component of this work in other works.}
\thanks{N. D. Riaziat, J. Chen, A. Krieger, and J. D. Brown are with the Johns Hopkins University Laboratory for Computational Sensing and Robotics, Baltimore, Maryland 21211, USA (\tt\small nriaziat@jhu.edu). 
}}

\begin{document}

\maketitle

\begin{abstract}

Electrosurgery is a surgical technique that can improve tissue cutting by reducing cutting force and bleeding. However, electrosurgery adds a risk of thermal injury to surrounding tissue. Expert surgeons estimate desirable cutting velocities based on experience but have no quantifiable reference to indicate if a particular velocity is optimal. Furthermore, prior demonstrations of autonomous electrosurgery have primarily used constant tool velocity, which is not robust to changes in electrosurgical tissue characteristics, power settings, or tool type. Thermal imaging feedback provides information that can be used to reduce thermal injury while balancing cutting force by controlling tool velocity. We introduce \gls{thermo} to autonomously reduce thermal injury while balancing cutting force by intelligently controlling tool velocity. We demonstrate \gls{thermo} in tissue phantoms and compare its performance to the constant velocity approach. Overall, \gls{thermo} improves cut success rate by a factor of three and can reduce peak cutting force by a factor of two. \gls{thermo} responds to varying environmental disturbances, reduces damage to tissue, and completes cutting tasks that would otherwise result in catastrophic failure for the constant velocity approach.

\end{abstract}


\section{Introduction}


\noindent Electrosurgery is a surgical technique for cutting tissue using energy from a high-frequency voltage source. Eight in ten surgical procedures use electrosurgery, often to remove diseased tissue~\cite{judith_lee_update_2008}. Monopolar electrosurgery uses a grounding pad to dissipate current through the body from a tool, directly heating the local tissue. However, electrosurgery can significantly damage nearby healthy tissue as well. Excess thermal damage can adversely affect surgery outcomes. Surgeons try to avoid generating thermal damage by modulating cut speed while ensuring cut accuracy. 

\begin{figure}[t]
    \centering
    \includegraphics[width=\linewidth]{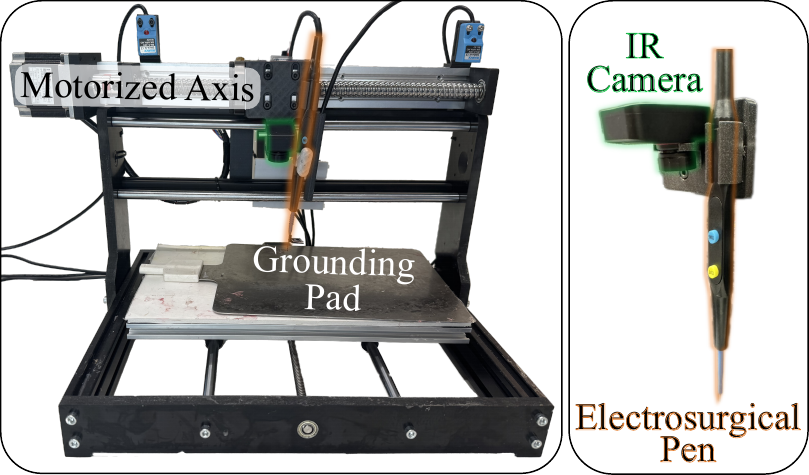}
    \caption{Single axis motorized testbed, holding a thermal camera, electrosurgical pen, and grounding plate. The testbed moves the electrocautery pen over a piece of phantom tissue to perform a linear cut. Speed can be controlled in real-time to test algorithm performance.}
    \label{fig:test_platform}
\end{figure}

When using electrosurgery to remove tissue, surgeons aim to spare as much healthy tissue as possible. Since signs of excess thermal energy are often not obvious until irreversible damage has been done, surgeons minimize dwell time by moving faster, as fast motions reduce the heat deposited in a specific area. However, fast motion comes at the cost of decreased cut accuracy and increased cutting force, which can also damage tissue. The ideal electrosurgical cut technique creates a small denatured tissue zone ahead of the tool tip that reduces mechanical stiffness and, thus, the required cut force. However, as the tool speed increases, this denatured tissue zone becomes smaller and eventually disappears, allowing tissue to accumulate on the tool tip and increasing the force needed to cut. This results in increased tissue damage, bulk tissue deformation, and lowered cut quality. Therefore, the optimal cutting velocity should maintain a balanced speed to reduce thermal damage and minimize excess force on the tissue.

While there have been no prior demonstrations of techniques for optimizing cut velocity in electrosurgery, there have been recent investigations into sensor-based automation for controlling thermal damage. Bao and Mazumder showed that thermal imaging can measure the denaturation zone and control it to a specific size using a novel computer-controllable electro \gls{esu}~\cite{bao_reduced_2022}. This method, however, contrasts with the fixed power level typically used by surgeons. El-Kebir et al.\ used thermal sensing and data-driven models to control the thermal damage boundary along a cut by pausing at discrete decision points~\cite{el-kebir_minimally_2023}. Unfortunately, neither approach simultaneously considers cut deformation and thermal damage to optimize the cut velocity. 

The same velocity modulation problem exists in autonomous robotic electrosurgery, which has had recent success in medical robotics research. While these autonomous approaches promise improved cut accuracy and decreased surgeon error, they have largely avoided real-time velocity optimization. Opfermann et al. ~\cite{opfermann_semi-autonomous_2017} demonstrated a visual servoing approach for electrosurgical cutting on the \gls{star}~\cite{shademan_supervised_2016}. Saeidi et al.\ similarly showed that STAR could perform tissue-cutting tasks using predetermined cut depth, power, and speed~\cite{saeidi_supervised_2019}. Ge et al.\ demonstrated tumor detection and excision using a suction gripper~\cite{ge_supervised_2021}. Each approach relies on a predetermined cut velocity, chosen based on clinical observation or simulation. Researchers have generally opted to move slowly to avoid excess tissue deformation. However, these approaches fail to account for the excess heat damage caused by slow cutting velocities.

Here, we introduce \gls{thermo}. \gls{thermo} uses thermal imaging to determine the optimal cut velocity by 1) dually identifying thermal and mechanical parameters online via a \gls{tukf} and 2) minimizing a parametrized cost function of denaturation width and cutting force. Thermal measurements inform the adaptation of thermal and mechanical parameters, which are applied to generic thermal and mechanical models to maximize cut accuracy and minimize thermal damage with respect to cut velocity. Collectively, we contribute \textcolor{red}{first steps towards}  1) \gls{ir}-camera-based tissue denaturation and force measurement method with accompanying validation, 2) a combined adaptive identification and optimization-based approach to control velocity based on thermal spread and cutting force, and 3) rigorous comparisons to the current cutting-edge autonomous electrosurgery approach.

\section{Methods and Materials}

\subsection{Electrocautery System}

The monopolar electrosurgery system consists of three main components: 1) the \gls{esu}, 2) the electrocautery pen, and 3) the grounding pad. The \gls{esu} supplies high-frequency voltage at a set power level to the tool. When a button on the tool is pressed, energy is conducted to the tool tip, through the tissue, and eventually to the grounding pad. The dispersion of this electrical energy to a large grounding area creates a high energy concentration near the tool tip, generating heat and the desired tissue denaturation/desiccation effect while safely dissipating low-concentration energy out of the patient's body. The following experiments utilize a DRE ASG-300 \gls{esu} (Avante Health Solutions, Concord, North Carolina) connected to a Medtronic Edge Electrosurgical Pencil, needle, and grounding pad. 

\subsection{Test Platform}

The single-axis motorized platform shown in Fig.~\ref{fig:test_platform} performs a electrosurgical linear tissue-cutting task. The platform moves the electrosurgical pen across a tissue sample at a commanded velocity. A NEMA 23 stepper motor drives a \SI{10}{\milli\meter\per rev}-pitch ball screw (FUYU FSL40, FUYU Technology, Chengdu, China) with \SI{250}{\milli\meter} stroke. A pulse train from an Arduino Minima microcontroller controls the step rate and, thus, the carriage speed. The thermal camera and tool are mounted to the carriage, so the tool is static in the thermal camera view. Tissue samples sit on the grounding pad underneath the linear stage for cutting. A serial interface to the Arduino delivers speed commands from the computer software.

\subsection{IR Imaging}
\label{sec:ir-imaging}

\begin{figure*}[t]
    \centering
    \includegraphics[width=\textwidth]{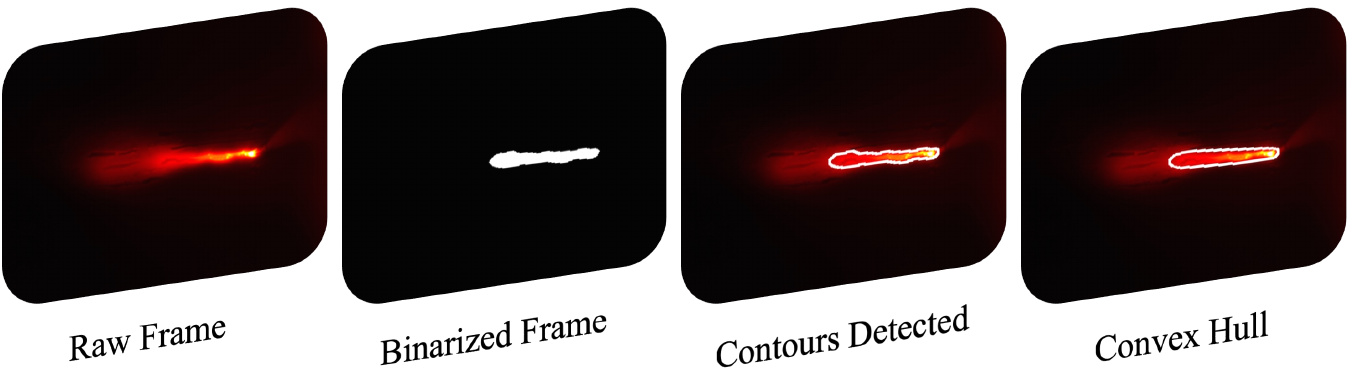}
    \caption{Frames from the \gls{ir} camera are binarized at the denaturation temperature. Contours are combined into one convex hull. The thermal width corresponds to the convex hull width.}
    \label{fig:cv}
\end{figure*}


The T3 Pro (IRay Technology Co., Ltd., Shanghai, China) captures a 384 x 288 pixel array at \SI{25}{\hertz} with a temperature lookup table to report absolute temperature values. To validate denaturation measurement accuracy, the camera's spatial accuracy, temporal accuracy, and temporal delay are characterized.

\subsubsection{Spatial Accuracy}

Accurate denaturation measurement is critical as minimizing excess tissue damage is a key clinical requirement. However, distortion in the image may prevent accurate spatial measurement. A pinhole model resolves this distortion and converts pixel width in the image frame to spatial width on the tissue surface. Calibration is conducted using the Zhang method in OpenCV~\cite{zhang_flexible_2000}, wherein a checkerboard pattern is used to find corners in the image and world frames. However, to create a checkerboard pattern compatible with thermal imaging, a black-and-white checkerboard pattern is illuminated by an \gls{ir}-emitting light source. Similar methods of calibration have been used by Shibata et al.~\cite{shibata_accurate_2017}



The denaturation boundary is found using methods originally proposed by Suzuki et al., implemented in OpenCV~\cite{suzuki_topological_1985}. A convex hull is approximated from the contours of the binarized denaturation boundary, and its width is reported as the denaturation width. The overall vision pipeline is shown in Fig.~\ref{fig:cv}.

Thirty nine unique images are used for camera calibration. With the pinhole calibration applied, the conversion ratio between the pixel and tissue surface is constant and was found using the same experimental setup as the pinhole calibration (\SI{4.81}{pixels\per\milli\meter}).


\subsubsection{Thermal Accuracy}

The temperature accuracy reported by the manufacturer is \SI{2}{\celsius} or 2\%. This is confirmed by measuring a black body source (Dahua HTBD, \SI{0.3}{\celsius}, Dahua Technology, Hangzhou, China) at \SI{50}{\celsius} from \SI{20}{\centi\meter}.



Of the 100 measurements of the black body source, the maximum error was \SI{1.4}{\celsius}, the mean error was \SI{0.86}{\celsius}, and the standard deviation was \SI{0.32}{\celsius}. 

\subsubsection{Temporal Delay}

Microbolometer-type thermal sensors respond to temperature changes according to the following first-order model~\cite{scaccabarozzi_about_2014}:

\begin{equation}
    \dot{T}(t) = \frac{1}{\tau} (T_\infty - T(t)),
\end{equation}

where $T$ is the measured temperature, $\tau$ is the first order time constant, and $T_\infty$ is the steady-state temperature. $\tau$ is determined by the thermal camera's 63.2\% rise time when exposed to a large temperature change. To determine $\tau$, the thermal camera is blocked by an insulating plate which is then rapidly removed, revealing the \SI{50}{\celsius} black body source. 

The thermal time constant $\tau$ was measured as \SI{17.6}{\milli\second}, which is consistent with the reported sampling frequency of this sensor. The thermal time constant for tissue is formulated as 
$\tau_{thermal} = \rho c_p V / h A_s$, where $\rho$ is the density, $c_p$ is the specific heat capacity, $V$ is the volume, $h$ is the heat transfer coefficient, and $A_s$ is the surface area. Taking values from the literature for human tongue, $\rho=\SI{1090}{\kilo\gram\per\meter^3}$, $c_p = \SI{3421}{\joule\per\kilo\gram\per\kelvin}$, and $h = \SI{3}{\watt\per\meter^2\kelvin}$~\cite{p_a_hasgall_itis_2022, kai_heat_2007}. Assuming a small section of tissue around the cut of \SI{1}{\milli\meter^3} volume and \SI{1}{\milli\meter^2} surface area, we recover $\tau_{thermal}=\SI{1254.4}{\second}$, much slower than the thermal response of our camera.

\subsection{Thermography for Electrosurgical Rate Management via Optimization}

\subsubsection{Force Model}

To maximize cut accuracy, cutting force must be limited. Tool deflection can be used as a proxy for cutting force using the following models. The tool position is modeled as a second order dynamical system:

\begin{equation}
    m\Ddot{x}_t = -k\,(x_t - x_n) + F_{cut},
    \label{eq:dynamics}
\end{equation}

\noindent where $x_t \in \mathbb{R}^2$ is the tool position, $k$ is the tool mechanical stiffness, $x_n \in \mathbb{R}^2$ is the neutral tool position when $F_{cut} = 0$. Prior research in thermal cutting has found the following inverse exponential model of cutting force~\cite{bain_thermomechanical_2011}:

\begin{equation}
    F_{cut} = -d(A\,e^{-bQ/u} + C),
\end{equation}
\noindent where $d$ is the depth of cut, $Q$ is the thermal heat input, $u$ is the cut velocity, and $A, b, C$ are cutting parameters. The cutting force formula is augmented to match experimental observations as follows:

\begin{equation}
    F_{cut}(u) = -d\,\exp(-C_{\text{defl}}/u) ,
    \label{eq:cutting_force}
\end{equation}

\noindent such that the exponential growth of force up to $d$ at a rate $C_{\text{defl}}$ with respect to velocity is accelerated at already high deflections. From Eq. \ref{eq:dynamics} thus

\begin{equation}
    \Ddot{x}_{t,i} = -\frac{1}{m}[k_t\,(x_{t,i} - x_n) -d\,\exp(-C_{\text{defl}}/u) ].
    \label{eq:tip_dynamics}
\end{equation}

Converting to a discrete-time state space model at timestep $i$, Eqs.~\ref{eq:tip_dynamics} and \ref{eq:cutting_force} become:

\begin{align}
    x_{i} &= \begin{bmatrix}
        x_{t, i} & \dot{x}_{t, i} & x_{n, i} & \hat{C}_{\text{defl}}
    \end{bmatrix}^T,  \label{eq:discrete_tip_dynamics_state} \\ 
    x_{i+1} &= f(x_i, u_i) + \mathcal{N}(0, Q),  \label{eq:discrete_tip_dynamics_predict} \\
    z_{i} &= \begin{bmatrix}
        x_{t, i} & x_{n, i}
    \end{bmatrix}^T + \mathcal{N}(0, R), \label{eq:discrete_tip_dynamics_measure} \\ 
\end{align}

\noindent where 

\begin{equation}
    f(x_i, u) = x_i + \Delta\,t\begin{bmatrix}
         \dot{x}_{t,i} \\
         -k (x_{t, i} - x_{n, i}) + F_{cut} \\
         \vec{0} \\ 
    \end{bmatrix}.
\end{equation}

The measurement model $z$ uses the right-most vertex of the convex hull to estimate the tool tip location. Then, the tool deflection is computed as $\delta = ||x_t - x_n||$. The parameters $k$ and $d$ are estimated experimentally and $\hat{C}_{\text{defl}}$ is estimated online. Parameter estimation is discussed in Sec.~\ref{sec:TUKF}.

\subsubsection{Thermal Model}

The electrosurgical task involves a moving heat source (the electrosurgical tool) in a heterogeneous material (the tissue). The planar problem is considered here, so the heat transfer problem with a 2D moving heat source is of the form 

\begin{equation}
    \pdv{T}{t} = \alpha ( \pdv[order=2]{T}{x} + \pdv[order=2]{T}{y}),
\end{equation}

where $\alpha$ is the thermal diffusivity, $T$ is the temperature field, and $t$ is time. It is assumed there is no heat transfer via radiation or convection, only conduction. 

If the moving heat source is moving in the x-axis, the coordinates may be transformed to the moving heat source frame by defining $\xi = x - u\,t$,
where $u$ is the heat source velocity. The following boundary conditions account for the moving heat source and the ambient conditions:

\begin{gather}
    \lim_{r\rightarrow0}2\pi r \lambda \pdv{T}{r} = Qu, \\
    \lim_{r\rightarrow\infty} T = T_{0},
\end{gather}

where $r = (\xi^2 + y^2)^\frac{1}{2}$ is the distance from the heat source, $\lambda$ is the tissue thermal conductivity, $Q$ is the heat flux from the moving heat source, and $u$ is the heat source velocity in the x-direction. As in Weichert et al.\ ~\cite{weichert_temperature_1978}, the ``quasi-stationary" assumption sets tool accelerations to zero and thus 

\begin{equation}
    T(\xi, r, u) = T_0(\xi, r) + \frac{Qu}{2\pi \lambda}\exp(-u\xi/2\alpha)K_0(ur/2\alpha),
    \label{eq:quasistationary}
\end{equation}

where $K_n(w)$ is a modified Bessel function of the second kind and $n$-th order. Eq.~\ref{eq:quasistationary} is used for the online feedback-based optimization strategy described in Section~\ref{sec:velopt}. 

Lu et al.\ expand on this model to directly predict an isotherm width~\cite{lu_width_2020}. In this application, the isotherm temperature is chosen to be the denaturation temperature, and thus the width refers to the denaturation width. The dimensional equation for the asymptotic isotherm width $w$ with correcting factor $f(T^*_c)$ is 

\begin{gather}  
    w(u) = \frac{1}{\sqrt{2\pi e}}\frac{q \alpha}{u \lambda d (T_c - T_o)}\,f(T^*_c), \label{eq:thermal_width} \\
    f(T^*_c) \approx \exp(-1/T^*_c)[1 + (1.477T^*_c)^{1.407}]^{0.7107}
\end{gather}

\noindent where the unit-less isotherm temperature $T^*_c = 2\pi\,\lambda\,d\,(T_c - T_0)q^{-1}$, $T_c$ is the isotherm temperature, and $\gamma = 0.5772$ is the Euler-Mascheroni constant. The correction factor $f(T^*_c)$ was designed by Lu et al.\ to account for velocity-related errors and regimes further from their asymptotic analysis. For the slow-moving regime, the maximum error after applying the correction factor is 10\%.

The tissue thermal conductivity $\lambda$ and density $\rho$ are estimated a priori, but the tissue specific heat capacity $c$ from $\alpha \coloneqq \lambda / \rho c$  and specific linear power density $\hat{q} \coloneqq q/d$ are unknown and may be varying throughout the sample. 

\subsubsection{Truncated Unscented Kalman Filter for Dual Estimation}
\label{sec:TUKF}

Given that many of the parameters in Eqs.~\ref{eq:discrete_tip_dynamics_predict}~and~\ref{eq:thermal_width} are unknown, we implement a dual state and parameter estimator using the unscented Kalman filter to account for our process non-linearities. The unscented Kalman filter (UKF) is a Kalman filter capable of applying non-linear prediction and measurement models using the so-called ``unscented transform," without Jacobian computation as the Extended Kalman Filter (EKF) requires. The joint state and parameter estimation is commonly solved using the non-linear Kalman filter. For example, Wan and van der Merwe showed the UKF for dual parameter and state estimation in their original introduction of the filter~\cite{wan_unscented_2000}. Further, Asadian et al.\ demonstrated the EKF to estimate needle insertion force and parameters~\cite{asadian_novel_2012}. The UKF estimates $\hat{\lambda}$, the augmented tissue thermal conductivity, $\hat{d}$, the augmented maximum cutting force, $\hat{q}$, the linear power density, and $\hat{c}$, the tissue specific heat capacity.

Consider a normally distributed $D-$dimensional variable $x_n$ which is subjected to a nonlinear function $f(.)$ to produce $y_n$:

\begin{gather}
 x_n \sim \mathcal{N}_D(\bar{\mathbf{x}}, \mathbf{P}_{xx}),    \\
 y_n = f(x_n).
\end{gather}

One way to estimate the density of $y_n$ is leveraging linear transforms to yield the new normal distribution. The unscented transform instead propagates a set of ``sigma" points $\{\mathbf{\mathcal{X}}_i\}^{2D}_{i=0}$ through the nonlinear function and estimates the resulting mean $\bar{\mathbf{y}}$ and covariance $\mathbf{P}_{yy}$.

\begin{gather}
    \mathbf{\mathcal{Y}}_i = f(\mathbf{\mathcal{X}}_i),\,\forall i\in {0, ..., 2D}, \label{eq:UT1} \\
    \mathbf{y}_i \sim \mathcal{N}(\bar{\mathbf{y}}, \mathbf{P}_{yy}).
\end{gather}

Normally, such a small sample could not accurately regenerate the population distribution; however, Julier and Uhlmann designed a strategic sigma point and weight selection method that samples the original distribution in a way that can capture the mean and covariance accurately up to the second-order. With the transformed sigma points $\mathbf{\mathcal{Y}}_i$ and weights $W_i$, the distribution's mean and covariance are:

\begin{gather}
    \bar{\mathbf{y}} = \sum_{i=0}^{2D} W_i \mathbf{\mathcal{Y}}_i \\
    \mathbf{P}_{yy} = \sum_{i=0}^{2D} W_i \{\mathbf{\mathcal{Y}}_i - \bar{\mathbf{y}}\}\{\mathbf{\mathcal{Y}}_i - \bar{\mathbf{y}}\}^T.
\end{gather}




One pitfall of this method, however, is the assumption of symmetric distribution. Thus, parameter values or their covariances may be physically infeasible (e.g., $\mathbb{R}^-$) when estimates are small and variances are comparatively large. The \gls{tukf} as proposed by Teixeira et al.\, constrains the parameter estimates. \gls{tukf} uses \gls{pdf} truncation on the posterior distribution $\rho(x_t | (z_t, \hdots, z_0))$. \gls{pdf} truncation generates a new mean state and covariance such that the distribution is defined only on the feasible interval. The truncation algorithm is detailed in~\cite{teixeira_state_2009}.

\begin{figure}[!t]
    \centering
    \input{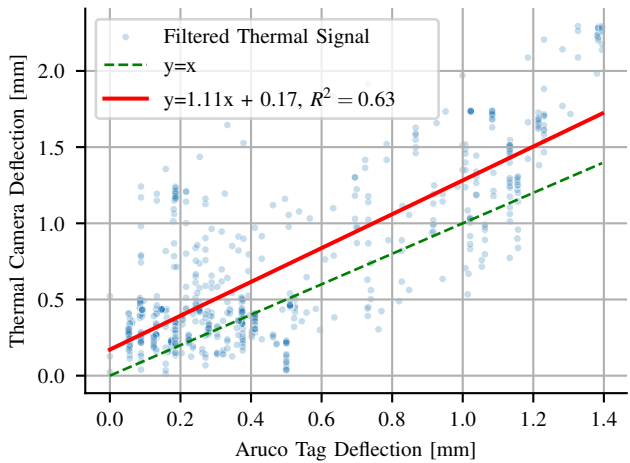}
    \caption{Comparison between deflection measurement using the thermal camera and the ArUco marker and endoscopic camera showed good agreement between the systems.}
    \label{fig:deflection_val}
\end{figure}

To validate the accuracy of this method, a small ArUco tag is mounted to the electrosurgical tool tip and measured from an endoscopic camera affixed to the tool holder. The estimated deflections from the thermal camera are compared to ground-truth results from the ArUco tag (see Fig.~ \ref{fig:deflection_val}). Root mean square tracking error is $\SI{0.23}{\milli\meter}$ and linearity is $R^2 = 0.63$.

\subsubsection{Velocity Optimization} \label{sec:velopt}

Two competing factors are clinically relevant to cut quality in electrosurgery: cell death area and tissue deformation. This section outlines a method to optimize both factors simultaneously. 

Consider the quadratic cost function

\begin{multline}
    J_i(v_i) = a F_i(v_i)^2 + b w_i(v_i)^2 + c (v_i - v_{i-1})^2 \\ +r(\min(v_i - \bar{v}, 0)),
    \label{eq:cost_func}
\end{multline}

\noindent where $F_i$ and $w_i$ are the tool force and thermal width at time $i$ respectively, as functions of the cut velocity $v_i$. $\bar{v}_i$ is the average surgeon's velocity during the cut. Parameters $a$, $b$, $c$, and $r$ are penalty weights on deflection, thermal width, input change, and ``time wasting," respectively. Time wasting is moving slower than the normal surgical procedure rate. This value should be minimized for better patient outcomes and lower medical costs. The cost $J$ is minimized at every time step by choosing the appropriate tool velocity. This form can be continuously differentiated with respect to $v$ to minimize $J$ to find the optimal move.

\subsection{Hydrogel Tissue Phantom}
\label{sec:Hydrogel}

Hydrogel phantoms are used as versatile, biocompatible substitutes for mimicking human tissue properties. Hydrogels are synthesized by mixing \gls{pva} with water. By adjusting the \gls{pva} concentration and freeze/thaw (F/T) cycles, their properties are tailored to replicate target tissues and are used in electrosurgical training~\cite{melnyk_mechanical_2020}. Compared to traditional animal models, hydrogel phantoms offer advantages regarding disposal, manufacturing ease, and consistency between models. Furthermore, unlike silicone-based alternatives, hydrogel phantoms can accommodate electrosurgical procedures, making them highly suitable for surgical research and training applications.

\begin{figure}[!t]
    \centering
    \includegraphics[width=\linewidth, trim=0 0 0 -0.7cm, clip]{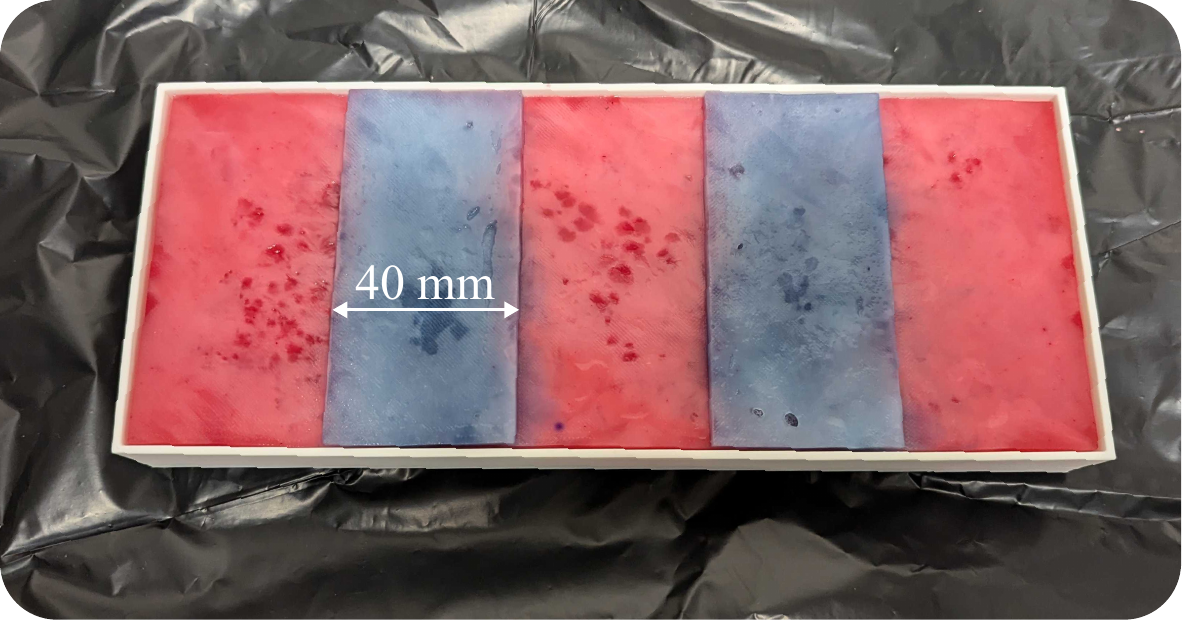}
    \caption{The hydrogel phantom is cast in a 3D printed mold with 2 or 3 mm steps along the length. Cotton fibers are laid into the blue (raised) sections to provide cutting resistance.}
    \label{fig:hydrogel}
\end{figure}

In our experiment, phantoms were created using a 10\% volume-to-weight \gls{pva} concentration. This solution was poured into a 3D-printed mold measuring \SI{250}{\milli\meter} in length and \SI{100}{\milli\meter} in width. The mold was divided into five \SI{50}{\milli\meter} segments, with the second and fourth regions raised and reinforced with cotton fibers to create variable cutting stiffness. Indentation heights of \SI{2}{\milli\meter} and \SI{3}{\milli\meter} are created for performance comparison. To enhance visual contrast and facilitate region identification during the experiments, the raised sections are blue while the remaining regions are red. The solution is subjected to a single freeze/thaw cycle at \SI{-20}{\celsius} to achieve solidification. A photograph of the completed hydrogel sample is shown in Fig.~\ref{fig:hydrogel}.

\section{Experiments and Results}

\begin{figure*}
    \centering
    \input{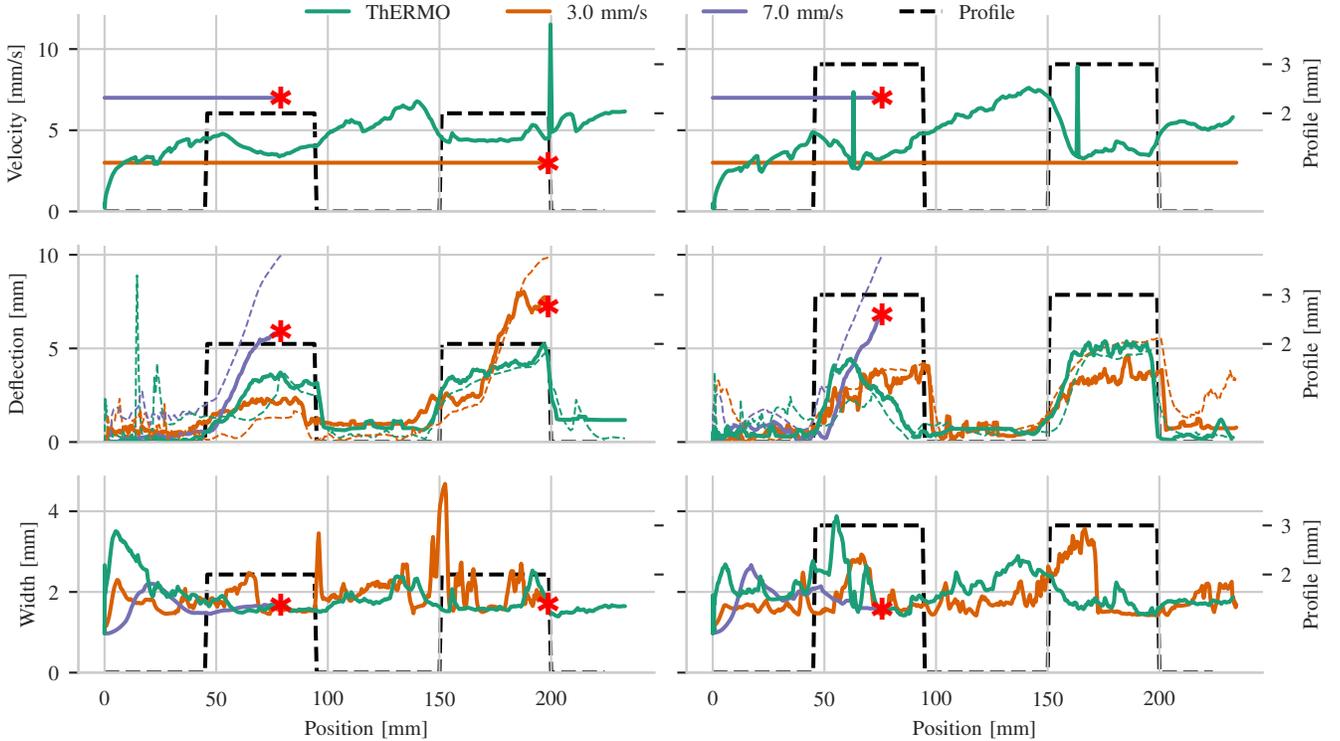}
    \caption{One trial each using \SI{3}{\milli\meter\per\second}, \SI{7}{\milli\meter\per\second}, and \gls{thermo} optimized cut velocity was conducted on hydrogel tissue phantoms with either a pair of 2 or \SI{3}{\milli\meter} raised steps. Columns, from left to right, \SI{2}{\milli\meter}, and \SI{3}{\milli\meter} steps (thick black dashed line). The top row shows velocity, the middle row shows deflection, and the bottom row shows thermal width. Colored dashed lines indicate deflection measured from the thermal camera, while solid lines represent ArUco tracker deflection measurements. Trials marked with a ``\textcolor{red}{\ding{81}}” failed prematurely.}
    \label{fig:metrics}
\end{figure*}

\begin{figure}[tb]
    \centering
    \input{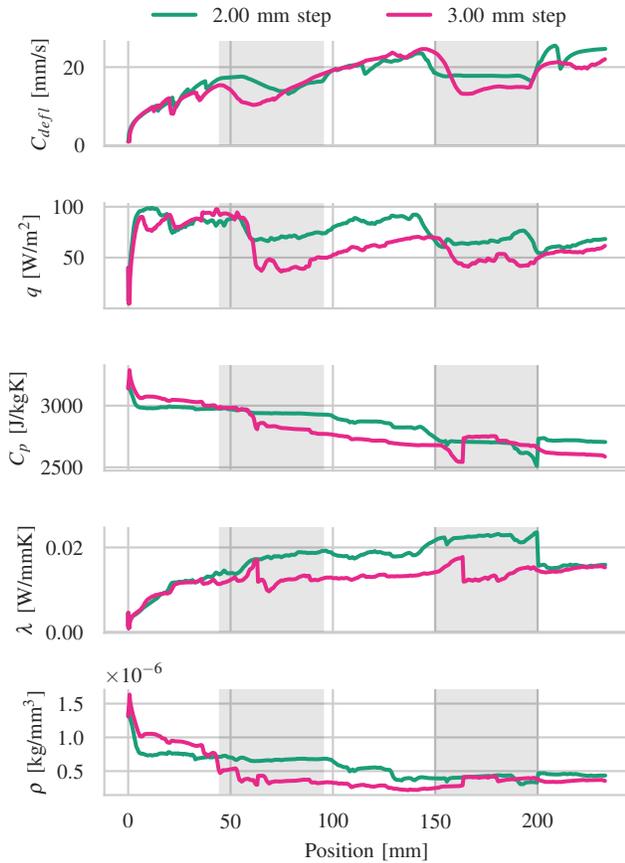}
    \caption{\gls{tukf} parameter estimates for 3 mm and 2 mm step cuts using \gls{thermo} are shown. Shaded regions denote the raised section. The parameters displayed are deflection rate, heat flux, specific heat capacity, thermal conductivity, and density from top to bottom. All parameters remain physically feasible and bounded.}
    \label{fig:parameters}
\end{figure}

\begin{figure}[tb]
    \centering
    \input{figures/combined.pgf}
    \caption{\gls{thermo} was compared to 3 mm/s and 7 mm/s cuts on 2 mm and 3 mm stepped phantoms. From left to right, the plots show unit-less disturbance rejection ratios for deflection and width and unitless success rate. The top row represents the 3 mm/s cut, the middle shows the 7 mm/s cut, and the bottom shows \gls{thermo}. Each condition was repeated six times. Lower values indicating better disturbance rejection.}
    \label{fig:success_step}
\end{figure}

\noindent The following experimental protocol aims to evaluate these objectives and comparisons to the constant velocity method: 1)  velocity selection stability, 2) parameter estimation stability, 3) thermal damage, 4) tool deflection, and 5) success rate.

Several successive \SI{200}{\milli\meter} linear cuts are performed using the constant velocity and \gls{thermo}. Clinical observation determines the constant velocity used in the naive approach. In hand-held ex-vivo tumor resection tasks, Ge et. al.\ measured an expert head and neck surgeon's average velocity to be \SI{7}{\milli\meter\per\second}, so experiments with the constant velocity method are run at this speed and \SI{3}{\milli\meter\per\second} for additional comparisons~\cite{ge_autonomous_2024} at slower velocities. Parameters used for Eq.~\ref{eq:cost_func} are shown in Tab. \ref{tab:weights} and were determined by equally valuing cutting force and thermal damage minimization while adding a small smoothing factor in $c$.

\subsection{Cutting Experiments}

\begin{table}[htb]
    \centering
    \caption{Experimental parameters used in Eq. \ref{eq:cost_func}.}
    \begin{tabular}{r c  c  c  c  c }
        \toprule
         \textbf{Parameter} & $a$ & $b$ & $c$ & $r$ & $\bar{v}$ \\
        \midrule
        \textbf{Value} & 1 & 1 & 0.001 & 0.001 & \SI{7}{\milli\meter\per\second}\\
        \bottomrule
    \end{tabular}
    \label{tab:weights}
\end{table}

The following procedure is used for each experiment. First, the \gls{esu} is set to \SI{30}{\watt}. Then, the stepped hydrogel tissue phantom (Sec.~\ref{sec:Hydrogel}) is placed on the grounding pad. The system is homed to the starting position, and the electrosurgical pen is lowered to \SI{2}{\milli\meter} below the contact point. \gls{thermo}, or the constant velocity method, is then used to cut the sample. The experiment is successfully completed when the tool reaches the end of the tissue or fails if the tool deflection exceeds \SI{10}{\milli\meter}. For the constant velocity experiments, if a \SI{7}{\milli\meter\per\second} cut fails, the experiment is repeated at \SI{3}{\milli\meter\per\second}. As many trials as possible are conducted with the fixed set of three hydrogel phantoms.

\subsubsection{Metrics}

The thermal margin is measured in situ by the thermal camera as the cut is performed. Measurements above \SI{60}{\celsius} are considered to signify tissue denaturation, where cells experience irreversible damage to their protein structures~\cite{sapareto_thermal_1984}. This measurement is validated based on the spatial, thermal, and temporal accuracy validations shown in Sec.~\ref{sec:ir-imaging}. The thermal margin should be minimized in width. Thermal widths are compared over the cut path. An ArUco tag and camera, as described in Section~\ref{sec:ir-imaging}, measure tool deflection independently of the thermal sensor. Tool deflection is compared throughout the cut. This metric reflects how well \gls{thermo} minimizes the cutting force and clinically indicates how spatially accurate the cut may be. The disturbance rejection ratio D is defined as $D = \frac{y_2 - y_1}{\Vert w\Vert}$, where $y_2$ is the mean output metric during the disturbance, $y_1$ is the mean output metric before the disturbance, and $\Vert w \Vert$ is the disturbance magnitude, in this case, the step change in profile height on the phantom tissue. Finally, the success rate is compared between conditions.

\subsection{Results}

In Fig.~\ref{fig:metrics}, the deflection and width results are shown for the three different conditions tested: a baseline condition with uniform tissue properties, a phantom with two \SI{2}{\milli\meter} steps, and a phantom with two \SI{3}{\milli\meter} steps. Each method is run three times on the baseline condition to show that \gls{thermo} performs at least as well as the constant velocity and finds the same rough steady-state velocity. \gls{thermo} displays similar or improved performance in both deflection and thermal width. Peak deflection is reduced by up to 50\% while maintaining a similar thermal width. Fig.~\ref{fig:parameters} shows the change in estimated parameters over a selected trial of \gls{thermo}. All parameters are bounded. Some parameters respond to the step change in parameters, while others are unchanged.

Fig.~\ref{fig:success_step} shows disturbance rejection ratios for deflection and width metrics. Width changes are similar between all conditions, but \SI{7}{\milli\meter\per\second} and \gls{thermo} outperform \SI{3}{\milli\meter\per\second} slightly. However, \gls{thermo} is nearly three times better for deflection in both scenarios.

Fig.~\ref{fig:success_step} shows the success rate for each of the nine scenarios. Both \SI{7}{\milli\meter\per\second} and \gls{thermo} successfully cut the hydrogel phantom when no disturbances were present. However, in \SI{2}{\milli\meter} step disturbance, \SI{7}{\milli\meter\per\second} succeeds 75\% of the time, and one additional trial at \SI{3}{\milli\meter\per\second} fails despite a 100\% success rate from \gls{thermo}. Only ~29\% of trials succeed with a \SI{7}{\milli\meter\per\second} cut in \SI{3}{\milli\meter} step phantoms. The trial at \SI{3}{\milli\meter} was successful. \gls{thermo} succeeded in ~85\% of trials.

\section{Discussion}
\label{sec:discussion}

\noindent \gls{thermo} performs superiorly to the naive approach in all experimental cases. Under varying tissue parameters, the parameter identification technique is stable and effective in tracking changing conditions. Not only is it demonstrated that the parameter estimates are stable and effective, but the improved outcomes also imply that the rate of adaptation is fast enough to track even step-like changing tissue characteristics. It is unclear if the parameter estimates reflect the true parameters of the tissue. The parameters converge to a set of valid but not necessarily true parameters. Though some sudden changes in parameters $q$, $C_{defl}$, and $\lambda$ are seen with respect to the disturbance region, the overall converging trend may indicate that the process noise in those parameters should be increased to allow for more rapid modulation. Future works may investigate parameter estimation accuracy.

The simplification of the 3D thermodynamics to a planar problem is justified since the z-dimensional heat transfer component is relatively small for thick tissue. Additionally, for this preliminary investigation, a 2D tissue surface is sufficient to show efficacy before moving to more complex 3D surfaces. 

The trials shown in Fig.~\ref{fig:metrics} show that constant velocity approaches often create large deflections when faced with stiffness parameter changes. Width changes were not as prevalent. This result is expected as the primary characteristics of the step and cloth change were mechanical, not thermal. \gls{thermo} does display some spikes in velocity, especially around the step edges. These are due to rapidly changing parameter estimates, but they effectively did not manifest physical performance changes on the testbed, perhaps due to the system inertia. Since each result represents a single trial (n=1), these performance measures should be interpreted as individual observations subject to sampling variation.

It may seem unintuitive that a trial at 3 mm/s should fail when some trials at 7 mm/s succeeded (Fig.~\ref{fig:success_step}); however, this result shows an additional detriment of slow traversal through tissue. As the electrode spends more time below the tissue surface, additional carbonization builds up on the needle tip, mechanically blunting it and reducing the electrical power transfer to surrounding tissue. The one failed trial of \gls{thermo} likely suffered from the same carbonization issue. 

Disturbance rejection ratios (Fig.~\ref{fig:success_step}) show equivalent, if not better, performance from \gls{thermo}. The step primarily represented a change in mechanical properties, so it is expected that thermal disturbance rejection ratios will be similar. The performance benefit is clear from the substantially improved mechanical disturbance rejection performance.

\section{Conclusions and Future Works}

\noindent \gls{thermo} performed effectively in phantom tissue experiments, minimizing thermal width and tool deflection by optimizing the cutting velocity. \gls{thermo} can also detect and prevent catastrophic errors when applied to autonomous electrosurgery. This is the first continuous-velocity optimization algorithm for electrosurgery. Potential application spaces for \gls{thermo} span handheld, laparoscopic, and robotic surgeries. The primary limitation will be engineering tool-mounted thermal cameras and finding case-specific methods for deflection or force measurement. 

Future work will consider the impact on real tissue and non-planar 3D cuts. The utility of such a system may be enhanced in a 3D cut, as cut depth may vary as a result of surface measurement errors. More accurate models of thermal cutting, of which there are few in the literature at this time, may also be developed. Further developments on \gls{thermo} may consider a finite-horizon optimization approach, such as Model Predictive Control (MPC), to improve optimality compared to the single-horizon greedy approach employed here. For applications in minimally invasive surgery, miniaturization of the thermal camera is necessary. Such efforts are ongoing to reduce the size of the sensor to fit within an endoscope~\cite{scutelnic_thermal_2022}. Finally, histological analysis may be valuable in confirming reduced cell-level thermal damage.

\FloatBarrier

\balance

\bibliographystyle{IEEEtran}
\bibliography{references.bib}

\end{document}